\documentclass[letterpaper]{article}
\usepackage[preprint]{aaai2027}
\usepackage[hyphens]{url}
\usepackage{graphicx}
\usepackage{natbib}
\usepackage{caption}
\usepackage{amsmath,amssymb}
\usepackage{booktabs}
\usepackage{array}
\usepackage{multirow}
\newcolumntype{C}[1]{>{\centering\arraybackslash}p{#1}}

\title{EP-LVMQA: Evidence-Preserving Coarse-to-Fine Localization for Low-Quality Long Video Moment Question Answering}
\author{Chen Guoning\textsuperscript{\rm 1}, Zhexun Hu\textsuperscript{\rm 1}, Yifan Wei\textsuperscript{\rm 2}, Yixuan Weng\textsuperscript{\rm 3}, Bin Li\textsuperscript{\rm 4}, Zhao Yufei\textsuperscript{\rm 1}}
\affiliations{
\textsuperscript{\rm 1}Nanyang Technological University, Singapore
\textsuperscript{\rm 2}Beihang University
\textsuperscript{\rm 3}Westlake University \\
\textsuperscript{\rm 4}Shenzhen Institutes of Advanced Technology, Chinese Academy of Sciences
}

\begin{document}
\maketitle

\begin{abstract}
Long instructional videos are increasingly used in medical education, industrial training, and procedural knowledge transfer, yet their evidence is often incomplete or unreliable: subtitles may be missing, ASR noisy, OCR sparse, and decisive actions briefly shown without narration. Existing VideoQA systems typically return labels or free-form text, making answer support difficult to inspect. We study Long Video Moment Question Answering (LVMQA), where a system returns timestamped answer-supporting evidence. We propose EP-LVMQA, an evidence-preserving coarse-to-fine framework comprising an Uncertainty-Aware Multimodal Evidence Timeline, a Coarse Evidence Planner, and an Evidence-Conditioned Fine Localizer. An evidence card transfers planner scores, matched cues, source-confidence proxies, modality-specific evidence, neighboring context, and a temporal prior between the latter two modules. On MedVidQA, EP-LVMQA achieves 26.1 R1@0.5 and 31.8 mIoU, improving the strongest non-ours results by 6.3 and 5.1 points. It raises Suff.@2 and Ver.@2 to 25.4 and 22.6, gains of 9.5 and 8.8 points. Under matched low-quality perturbations, average relative degradation is reduced from the strongest non-ours value of 14.6\% to 11.1\%. On Ego4D-NLQ and QVHighlights, EP-LVMQA remains broadly comparable to established task-specific localization methods.
\end{abstract}

\section{Introduction}
Video question answering (VideoQA) increasingly couples answer prediction with temporal grounding \cite{jang2017tgifqa,lei2018tvqa,xiao2024trust,di2024groundvqa}. Natural-language video localization has likewise progressed from proposal-based matching to set prediction and unified grounding \cite{gao2017tall,lei2021qvhighlights,moon2023qddetr,lin2023univtg}. Yet most systems either generate an answer or retrieve a query-matching interval; they do not preserve the multimodal provenance needed to inspect whether the interval actually supports the answer.

\begin{figure*}[t]
\centering
\includegraphics[width=0.95\textwidth]{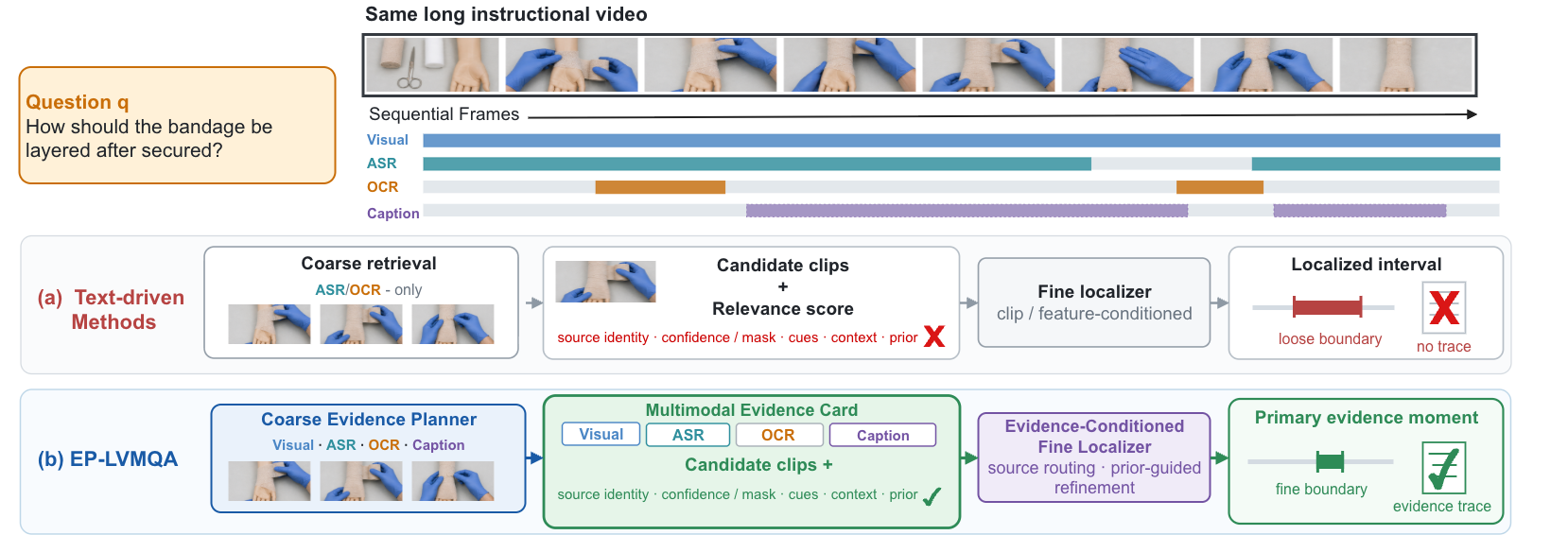}
\caption{(a) Text-driven methods pass ASR/OCR-retrieved clips or scores to a clip-conditioned localizer, yielding a loose interval without provenance. (b) EP-LVMQA transfers multimodal evidence and source-aware metadata in an evidence card, enabling routed, prior-guided refinement and an inspectable primary evidence moment.}
\label{fig:overview}
\end{figure*}

This gap matters in long instructional videos, where answer evidence spans unreliable sources. MedVidQA pairs medical questions with timestamped visual answers \cite{gupta2023medvidqa}, but real recordings may contain noisy ASR, sparse OCR, and plausible yet visually unsupported captions. Answer-seeking questions may also express needs that do not literally name the target action. Long-video methods reduce search cost through window selection and coarse-to-fine alignment \cite{hou2023cone,pan2023scanning,barrios2023guidance,mu2024snag}. Figure~\ref{fig:overview}(a) abstracts the text-driven variant: ASR or OCR guides coarse retrieval, but only candidate clips, features, or relevance scores reach the fine localizer. This handoff discards source identity, confidence, availability masks, matched cues, neighboring context, and the temporal prior.

The discarded information affects localization quality, not merely the explanation produced afterward. A generated caption can make a broadly related window rank highly even when a brief visual action in a neighboring window provides the decisive evidence. A clip-conditioned localizer cannot distinguish mutually supporting sources from a single unreliable cue or recover the planner's temporal rationale. The retrieval-to-localization interface therefore determines whether the model can reject misleading evidence, refine both boundaries, and expose why the returned moment supports the answer.

We therefore study \emph{Long Video Moment Question Answering} (LVMQA): given a long video and an answer-seeking question, return a timestamped primary evidence moment rather than unconstrained text. Figure~\ref{fig:overview} contrasts the two interfaces. Text-driven methods pass candidate clips and relevance scores to a clip-conditioned localizer, often yielding a loose interval without an inspectable trace. EP-LVMQA instead transfers a multimodal evidence card with source-specific payloads and retrieval metadata, enabling routing and prior-guided refinement. The target is a finely bounded primary evidence moment accompanied by an evidence trace.

\textbf{EP-LVMQA} operationalizes this design with an Uncertainty-Aware Multimodal Evidence Timeline, a Coarse Evidence Planner, and an Evidence-Conditioned Fine Localizer. The card transfers the planner score, modality payloads, confidence proxies, availability masks, matched cues, neighboring context, and a temporal prior. Its retained fields also support the returned evidence trace.

Our contributions are threefold: (1) we formulate low-quality instructional VideoQA as localization of inspectable answer evidence; (2) we introduce an Evidence-Preserving Candidate Interface with source-aware routing and prior-guided refinement; and (3) on MedVidQA, controlled perturbations, and two cross-benchmark checks, we show gains in localization, robustness, and human-audited evidence quality while identifying planner misses and imperfect faithfulness as limitations.

\section{Related Work}
\textbf{Grounded VideoQA.} VideoQA links answers to spatio-temporal evidence \cite{jang2017tgifqa,lei2018tvqa,xiao2024trust,di2024groundvqa}. Instructional grounding spans TACoS and HowTo100M \cite{regneri2013grounding,miech2019howto100m}. EP-LVMQA instead outputs timestamped evidence with provenance.

\textbf{Temporal grounding.} Methods include proposal/span prediction \cite{gao2017tall,hendricks2017localizing,ghosh2019excl,zhang2020vslnet}, 2D maps \cite{zhang2020t2tan}, DETR localization \cite{lei2021qvhighlights,moon2023cgdetr,moon2023qddetr}, joint retrieval/highlights \cite{liu2022umt,lin2023univtg}, and long-video grounding \cite{hou2023cone,pan2023scanning,mu2024snag,yuan2025momentseeker}. Movie grounding and Video-LLMs extend duration \cite{soldan2022mad,huang2024vtimellm,wang2025grounded}. We instead preserve evidence across stages.

\textbf{Long-video evidence selection.} Sparse selection can erase source identity. Pretraining, missing-modality modeling, calibration, and diversity reranking improve reliability \cite{alayrac2020selfsupervised,bain2021frozen,neverova2016moddrop,ma2022missing,guo2017calibration,carbonell1998mmr}. Ego4D methods reason over localized windows \cite{grauman2022ego4d,hou2023groundnlq}. EP-LVMQA exposes an auditable moment; MomentSeeker is only an external MedVidQA planner baseline.

\section{Task Formulation}
Given $\mathcal{D}=\{(V_i,q_i,M_i^\ast)\}_{i=1}^{N}$, where $M_i^\ast$ contains annotated answer-supporting intervals, the model predicts
\begin{equation}
\widehat M=\{(\hat s_k,\hat e_k,\hat c_k,\hat r_k)\}_{k=1}^{N_b}.
\end{equation}
Here $(\hat s_k,\hat e_k)$ is an evidence moment, $\hat c_k$ its confidence, and $\hat r_k$ an externally inspectable provenance record rather than a hidden reasoning chain. Although multiple intervals are permitted, we evaluate the top-ranked prediction against the best-overlapping annotation. At temporal unit $t$, we retain
\begin{equation}
y_t=\{v_t,x_t^{\mathrm{asr}},x_t^{\mathrm{ocr}},x_t^{\mathrm{cap}},m_t,c_t\},
\end{equation}
where $m_t$ and $c_t$ encode source availability and confidence. Generated captions are uncertain cues, not ground-truth subtitles.

The Evidence-Preserving Candidate Interface transfers a model-facing \emph{evidence card} from the planner to the localizer. We distinguish this card from the returned \emph{evidence moment} and its output-facing \emph{evidence trace}. This distinction prevents the interface from being reduced to a post-hoc explanation.

\section{Method}
\begin{figure*}[t]
\centering
\includegraphics[width=\textwidth]{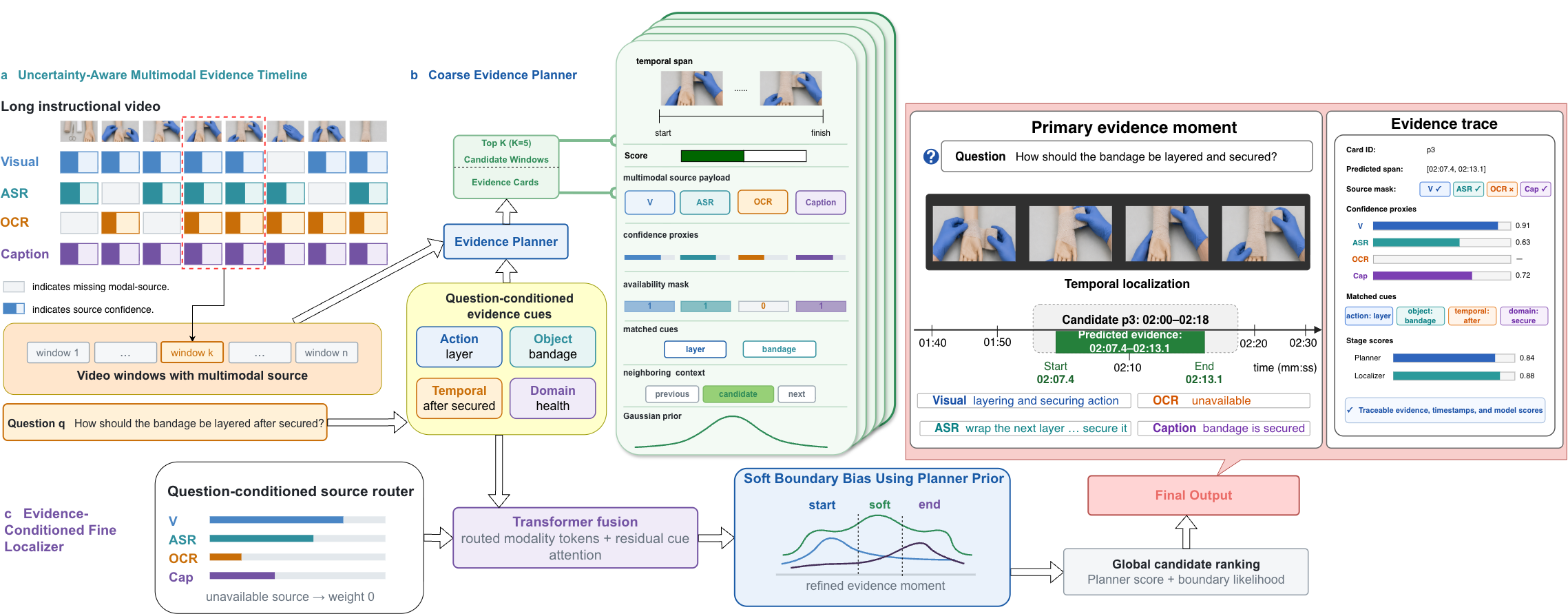}
\caption{EP-LVMQA comprises three modules: (a) an Uncertainty-Aware Multimodal Evidence Timeline, (b) a Coarse Evidence Planner, and (c) an Evidence-Conditioned Fine Localizer. Question-Conditioned Evidence Cues condition the planner and localizer, which communicate through an Evidence-Preserving Candidate Interface implemented as an evidence card.}
\label{fig:architecture}
\end{figure*}

Figure~\ref{fig:architecture} organizes EP-LVMQA into three modules. Question-Conditioned Evidence Cues condition the planner and localizer, which exchange an evidence card through the Evidence-Preserving Candidate Interface.

\subsection{Uncertainty-Aware Multimodal Evidence Timeline}
Visual features, ASR, OCR, and generated captions are aligned into overlapping windows without collapsing source identity. For modality $m$ in window $j$, a label-free confidence proxy combines availability, calibrated native confidence, and local cross-modal agreement:
\begin{equation}
\rho_j^m=\frac{\lambda_a A_j^m+\lambda_c u_j^m C_j^m+\lambda_gG_j^m}
{\lambda_a+\lambda_cu_j^m+\lambda_g}.
\end{equation}
We use $(\lambda_a,\lambda_c,\lambda_g)=(0.4,0.3,0.3)$ and set $\rho_j^m=0$ when unavailable. Caption and visual streams have no native confidence. This proxy guides planning and routing but is not claimed to estimate true reliability.

Native ASR/OCR confidence is clipped at the 1st and 99th training percentiles and min--max calibrated; $G_j^m$ is mean projected cosine agreement with available streams within $\pm4$\,s. The planner uses the masked mean confidence, while the localizer learns question-dependent source weights. Separating these operations avoids using the localizer's learned routing score as evidence for its own candidate selection.

\subsection{Coarse Evidence Planner}
The planner compresses the search space while preserving candidate recall. Each window retains its span, source features, mask, and mean confidence.

\paragraph{Question-Conditioned Evidence Cues.}
To bridge indirect questions to observable actions, a fixed rule-assisted parser extracts action/verb-particle, object, temporal, and high-IDF domain cues, retaining at most four per type. A cue survives only if its maximum text/visual/caption similarity to memory exceeds $\theta_c=0.30$; otherwise only the question embedding is used. The planner scores a window as
\begin{equation}
S_j=f_\theta([h_q;h_j^{\rm vis};h_j^{\rm asr};h_j^{\rm ocr};
h_j^{\rm cap};\bar\rho_j;m_j;\psi_j]),
\end{equation}
where $\psi_j$ encodes position, duration, and neighboring agreement. MMR selection ($\lambda_{\rm MMR}=0.20$) reduces redundancy; adjacent windows within 4\,s are merged subject to a 96\,s cap and a global duration budget.

The planner targets evidence recall rather than boundary accuracy. Its compact candidate union should cover the decisive moment; candidate duration and an oracle upper bound separate search from localization failure.

\paragraph{Evidence-Preserving Candidate Interface.}
The interface instantiates each selected candidate $j$ as an evidence card
\begin{equation}
p_j=\{\tau_j,S_j,\Omega_j,\rho_j,m_j,\xi_j,\nu_j,\pi_j\},
\end{equation}
containing its span, score, modality payload, confidence and mask, matched cues, neighboring context, and a Gaussian temporal prior. Training labels and human audit scores are excluded from model input.

The prior is centered on the highest-scoring constituent window, with standard deviation equal to half the window length. Immediate neighboring summaries help distinguish repeated steps without restoring the full video. Model-input fields are versioned separately from training-only IoU labels and audit-only judgments.

\subsection{Evidence-Conditioned Fine Localizer}
The localizer encodes card metadata, question tokens, and modality tokens, then refines each card into a bounded evidence moment. A question-conditioned router assigns
\begin{equation}
\omega_j^m=\frac{m_j^m\exp(W_r[h_q;h_j^m;\rho_j^m])}
{\sum_{m'}m_j^{m'}\exp(W_r[h_q;h_j^{m'};\rho_j^{m'}])+\epsilon},
\end{equation}
so unavailable sources receive zero weight. A transformer fuses routed tokens and residual cue attention. Start and end distributions use the planner prior as a soft boundary bias:
\begin{align}
P_s^j(t)&=\mathrm{softmax}(W_sH_j(t)+\eta\log\pi_j(t)),\\
P_e^j(t)&=\mathrm{softmax}(W_eH_j(t)+\eta\log\pi_j(t)),
\end{align}
with $\eta=0.5$. Candidate predictions are ranked by planner score plus start/end log-likelihood; the top interval is returned with its evidence trace.

More precisely, valid boundaries maximize $P_s^j(s)P_e^j(e)$ under $s\leq e$ and the 96\,s cap. Global ranking uses
\begin{equation}
\ell_j=S_j+\log P_s^j(\hat s_j)+\log P_e^j(\hat e_j).
\end{equation}
The trace records the card identifier, timestamps, cues, source masks, confidence proxies, and planner/localizer scores. These fields support inspection but do not prove causal reliance.

\subsection{Training}
Training is sequential. Planner windows are positive at maximum IoU$\geq0.3$, negative below 0.1, and otherwise ignored. Binary cross-entropy is combined with margin ranking (margin 0.2) against semantically similar non-overlapping negatives; the selected planner is then frozen and Top-$K$ candidates are cached.

The localizer minimizes start/end cross-entropy plus IoU loss. Over available modalities only, normalized confidence $\widetilde\rho_j$ gives a weak consistency term
\begin{equation}
\mathcal{L}_{\rm fine}=\mathcal{L}_{\rm loc}
+0.05D_{\rm KL}(\widetilde\rho_j\Vert\omega_j).
\end{equation}
This term is a routing prior, not reliability supervision. There is no gradient through hard candidate selection, and no ground-truth timestamp, answer label, or audit judgment enters confidence computation or test-time cards.

\section{Experiments}
\subsection{Setup}
\label{sec:experimental-setup}
\textbf{Datasets and metrics.} MedVidQA is the target benchmark; Ego4D-NLQ and QVHighlights test cross-domain grounding compatibility \cite{gupta2023medvidqa,grauman2022ego4d,lei2021qvhighlights}. We report rank-1 recall at temporal IoU thresholds, mIoU, planner recall, candidate duration, compression, and runtime. Baselines include VSLNet, Moment-DETR, QD/CG-DETR, UniVTG, CausalVTG, SRF, CONE, GroundNLQ, Grounded-VideoLLM, and TimeZero \cite{zhang2020vslnet,lei2021qvhighlights,moon2023qddetr,moon2023cgdetr,lin2023univtg,wang2025causalvtg,lee2026srf,hou2023cone,hou2023groundnlq,wang2025grounded,wang2025timezero}. Reproduced feature-based methods use matched preprocessing and budgets; Video-LLMs use matched prompts and evidence budgets. Official results are reference-only.

For localization, $\Rone@\tau$ is the fraction of questions whose top-ranked interval overlaps any annotation by at least $\tau$. Planner recall requires at least one selected candidate at IoU$\geq0.3$. Candidate duration is the duration of the union of selected candidates divided by video duration, and compression is its reciprocal. The long-duration split contains official MedVidQA test videos longer than 900\,s and all associated QA pairs.

\textbf{Human-audit design.} We fixed 300 MedVidQA queries before inspecting system outputs. Input-only strata combine video-duration quartile, question-complexity bin, and source availability/quality; model scores and predictions never determine inclusion. Every method is evaluated on the same queries. Method identity and output order are concealed from three raters. The question and reference answer or decision target are shown, but the method, evidence trace, and ground-truth timestamps are hidden. The keep view contains the predicted moment with a 2-s playback margin on each side and visibly marks the original boundaries.

Sufficiency and Verifiability are the two primary human outcomes. For Sufficiency, 0 means absent or misleading answer support, 1 partial but incomplete support, and 2 evidence sufficient to answer the question. For Verifiability, 0 means no directly identifiable answer evidence, 1 relevant evidence with incomplete context or imprecise boundaries, and 2 directly inspectable evidence with appropriate temporal scope. Sufficiency therefore targets semantic completeness, whereas Verifiability targets temporal inspectability. A 30-item pilot fixes the rubric before formal scoring. For seed $s$, query $i$, criterion $c\in\{\mathrm{Suff},\mathrm{Ver}\}$, and rater $r$,
\begin{align}
I_{is}^{c@2}&=\mathbb{I}\!\left[\sum_{r=1}^{3}
\mathbb{I}(y_{isr}^{c}=2)\geq2\right],\\
c@2&=\frac{100}{S}\sum_{s=1}^{S}\frac{1}{N_s}\sum_i I_{is}^{c@2}.
\end{align}
Here $S=3$, and $N_s$ is the effective number of audited queries for seed $s$. Necessity uses the duration-preserving drop view and is retained only as an auxiliary irreplaceability diagnostic because equivalent evidence may remain elsewhere. The supplement specifies the display sequence, repeated-evidence flag, full rubric, adjudication, agreement, seed aggregation, paired uncertainty, multiplicity family, and duration-adjusted construct checks. Main-table audit values are measured seed-mean point estimates; inferential claims require the corresponding raw-label analysis.

\textbf{Fairness protocol.} ``Matched'' feature-based baselines share temporal preprocessing, comparable source views and budgets, and timestamp post-processing. ``Matched-prompt'' Video-LLMs receive the same frame-token budget and aligned ASR/OCR/caption view but no evidence card, planner candidates, source router, or audit metadata. Architectures retain their original encoders and fusion modules. Table~\ref{tab:main-results} marks each row as reported/official (Off.), independently reproduced (Rep.), matched feature-based (M), matched-prompt (MP), or ours. Official rows are reference-only and are excluded from strict budget-matched superiority claims.

\textbf{Implementation.} Videos are decoded at 1 fps. Frozen SlowFast and CLIP features represent video and text \cite{feichtenhofer2019slowfast,radford2021clip}; Whisper large-v3, PP-OCRv4, and LLaVA-OneVision-7B provide ASR, OCR, and 4-s factual captions \cite{radford2023whisper,du2020ppocr,li2024llavaonevision}. Windows are 16\,s with 8\,s stride (32/16\,s for videos over 15 minutes). The default is $K=5$ and
\begin{equation}
B_{\rm dur}=\min(0.15|V|,300\,\mathrm{s}).
\end{equation}
The planner is a two-layer 512-D MLP; the localizer is a four-layer 512-D transformer. AdamW trains the planner/localizer for at most 30/50 epochs with learning rates $2{\times}10^{-4}/10^{-4}$ and validation early stopping.

SlowFast and CLIP yield 2816-D visual tokens projected to 512 dimensions. Whisper uses greedy word timestamps; OCR retains English detections above 0.50; the frozen captioner describes visible action, manipulated object, and procedural state in at most 32 new tokens. Candidates use at most 128 temporal tokens. Both stages use weight decay $10^{-4}$, gradient clipping at 1.0, 5\% warmup, and seeds $\{0,1,2\}$.

\subsection{Main Results}
\begin{table*}[t]
\centering
\small
\setlength{\tabcolsep}{0pt}
\begin{tabular}{@{}p{0.169\textwidth}@{}C{0.070\textwidth}@{}*{10}{C{0.0735\textwidth}@{}}}
\toprule
\multirow{3}{*}{Method} & \multirow{3}{*}{Prot.} & \multicolumn{6}{c}{MedVidQA} & \multicolumn{2}{c}{Ego4D-NLQ} & \multicolumn{2}{c}{QVHighlights} \\
\cmidrule(lr){3-8}\cmidrule(lr){9-10}\cmidrule(lr){11-12}
& & \multicolumn{3}{c}{R1} & \multirow{2}{*}{mIoU} & \multirow{2}{*}{Suff.@2} & \multirow{2}{*}{Ver.@2} & \multicolumn{2}{c}{R1} & \multirow{2}{*}{R1@0.5} & \multirow{2}{*}{MR-mAP} \\
\cmidrule(lr){3-5}\cmidrule(lr){9-10}
& & @0.3 & @0.5 & @0.7 & & & & @0.3 & @0.5 & & \\
\midrule
MedVidQA/MVAL & Off./Rep. & 25.8 & 14.2 & 6.5 & 20.1 & 10.2 & 8.8 & -- & -- & -- & -- \\
VPTSL/TAGV-style & Rep. & 32.6 & 18.2 & 8.4 & 25.8 & \underline{15.9} & \underline{13.8} & -- & -- & -- & -- \\
Moment-DETR \cite{lei2021qvhighlights} & M & 25.7 & 14.1 & 6.3 & 20.0 & 10.0 & 8.6 & 4.57 & 1.85 & 59.32 & 35.71 \\
CG-DETR \cite{moon2023cgdetr} & M & 32.1 & 18.4 & 9.3 & 24.5 & 13.6 & 11.6 & -- & -- & \textbf{67.91} & \underline{47.16} \\
UniVTG w/ PT \cite{lin2023univtg} & M & 34.3 & \underline{19.8} & 10.1 & 26.0 & 14.8 & 12.6 & 11.74 & 7.54 & 64.82 & 42.95 \\
CausalVTG \cite{wang2025causalvtg} & M & 34.0 & 19.2 & \underline{10.6} & 25.7 & 14.9 & 12.9 & 19.12 & 13.41 & 66.05 & \textbf{47.50} \\
SRF \cite{lee2026srf} & M & 34.8 & 19.6 & 10.5 & 26.3 & 15.4 & 13.5 & 17.98 & 11.53 & 67.31 & 43.97 \\
GroundNLQ \cite{hou2023groundnlq} & Off. & -- & -- & -- & -- & -- & -- & \textbf{25.67} & \textbf{18.18} & -- & -- \\
Grounded-VideoLLM \cite{wang2025grounded} & MP & 33.9 & 18.7 & 7.5 & 25.4 & 12.9 & 10.8 & 14.12 & 8.35 & 56.21 & 33.42 \\
TimeZero \cite{wang2025timezero} & MP & \underline{36.4} & 19.3 & 7.6 & \underline{26.7} & 13.8 & 11.5 & 15.92 & 9.79 & 58.70 & 36.21 \\
\midrule
\textbf{EP-LVMQA} & \textbf{Ours} & \textbf{41.2} & \textbf{26.1} & \textbf{14.3} & \textbf{31.8} & \textbf{25.4} & \textbf{22.6} & \underline{21.83} & \underline{13.60} & \underline{67.42} & 45.69 \\
\bottomrule
\end{tabular}
\caption{Main comparison with explicit result provenance. MedVidQA is the target setting; Ego4D-NLQ and QVHighlights are compatibility checks. Prot.: Off.=reported/official reference, Rep.=our reproduction, M=matched feature-based evaluation, MP=matched-prompt Video-LLM evaluation. Off./Rep. means that localization is official while audit values come from reproduced saved predictions; these values are not combined into a matched superiority claim. All audit columns are measured strict-majority rates on the common blinded 300-query set. Bold and underlining denote the best and second-best listed value per metric, irrespective of protocol class.}
\label{tab:main-results}
\end{table*}

\begin{figure}[t]
\centering
\includegraphics[width=\columnwidth]{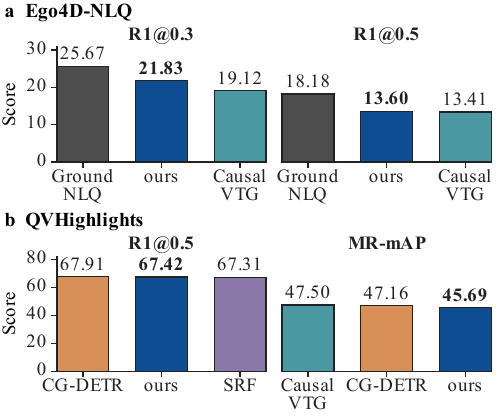}
\caption{Cross-benchmark performance on (a) Ego4D-NLQ and (b) QVHighlights, showing the top three methods for each metric. \emph{Ours} denotes EP-LVMQA.}
\label{fig:cross-benchmark}
\end{figure}

EP-LVMQA leads every MedVidQA metric in Table~\ref{tab:main-results}. Relative to the strongest listed non-ours value, its gains are 13.2\%, 31.8\%, 34.9\%, and 19.1\% for $\Rone@0.3$, $\Rone@0.5$, $\Rone@0.7$, and mIoU, respectively. Beyond the target setting, EP-LVMQA shows reasonable cross-benchmark compatibility without benchmark-specific design, while performing only slightly below specialized methods, as shown in Figure~\ref{fig:cross-benchmark}.

\noindent\textbf{Finding 1: EP-LVMQA exhibits larger gains at stricter IoU thresholds.}
Against matched UniVTG, the relative gain rises from 20.1\% at $\Rone@0.3$ to 31.8\% at $\Rone@0.5$ and 41.6\% at $\Rone@0.7$. The widening margin at stricter IoU thresholds indicates that EP-LVMQA's gains arise not only from coarse retrieval but also from more precise boundary localization enabled by the Evidence-Preserving Candidate Interface, allowing it to better isolate answer-bearing actions from adjacent procedural steps.

\noindent\textbf{Finding 2: Human-audit metrics capture practical utility beyond conventional localization scores.}
EP-LVMQA improves the strongest listed non-ours Suff.@2 and Ver.@2 values by 59.7\% and 63.8\%, exceeding its relative gains on IoU-based metrics. The strongest localization and audit baselines are also different models. Sufficiency tests whether the returned content answers the question, whereas Verifiability additionally requires identifiable and well-scoped evidence. A prediction can overlap the annotation yet remain too broad, incomplete, or ambiguous for either criterion. Temporal overlap is therefore related to, but not equivalent to, practical answer support. Nevertheless, the absolute rates remain low---only 25.4\% and 22.6\%, respectively---and are still insufficient to support reliable use in real-world human-facing settings.

\subsection{Robustness and Planner Analysis}
\begin{table*}[!t]
\centering
\small
\begin{tabular}{@{}p{0.22\textwidth}@{}*{7}{C{0.108\textwidth}@{}}}
\toprule
Method & Clean & ASR-50 & OCR-50 & HN-Cap & Conflict & Avg. drop $\downarrow$ & Conflict Suff.@2 \\
\midrule
VPTSL/TAGV-style & 18.2 & 11.4 & 16.6 & 13.2 & 12.0 & 26.9 & 9.1 \\
Moment-DETR \cite{lei2021qvhighlights} & 14.1 & 11.5 & 13.1 & 10.9 & 10.2 & 19.0 & 6.8 \\
QD-DETR \cite{moon2023qddetr} & 16.7 & 13.4 & 15.4 & 12.5 & 11.7 & 20.7 & 7.9 \\
CG-DETR \cite{moon2023cgdetr} & 18.4 & 15.2 & 16.9 & 13.7 & 13.0 & 20.1 & 9.0 \\
UniVTG \cite{lin2023univtg} & \underline{19.8} & \underline{16.7} & 18.2 & 15.0 & 14.4 & 18.8 & 10.6 \\
CausalVTG \cite{wang2025causalvtg} & 19.2 & 16.5 & 17.9 & \underline{15.8} & \underline{15.4} & \underline{14.6} & \underline{11.9} \\
SRF \cite{lee2026srf} & 19.6 & 16.6 & \underline{18.3} & 14.7 & 13.3 & 19.8 & 10.2 \\
\textbf{EP-LVMQA} & \textbf{26.1} & \textbf{23.7} & \textbf{24.8} & \textbf{22.3} & \textbf{22.0} & \textbf{11.1} & \textbf{20.8} \\
\bottomrule
\end{tabular}
\caption{MedVidQA robustness in R1@0.5. ASR/OCR-50 removes half the corresponding evidence; HN-Cap uses semantically similar non-overlapping caption negatives; Conflict retains ASR but replaces captions. Avg. drop is the mean relative loss across the four perturbations. All methods receive the same degraded input views.}
\label{tab:robustness}
\end{table*}

EP-LVMQA has the highest score under every perturbation and the lowest mean relative drop, 11.1\%, compared with the strongest non-ours value of 14.6\% achieved by CausalVTG (Table~\ref{tab:robustness}). This is a 3.5-point absolute and 24.0\% relative reduction in average degradation.

\noindent\textbf{Finding 3: EP-LVMQA's advantage widens under semantic distraction and cross-modal conflict.}
Relative to the strongest non-ours method in each column, EP-LVMQA gains 41.9\% under ASR-50 over UniVTG and 35.5\% under OCR-50 over SRF. Under incorrect-evidence perturbations, it outperforms CausalVTG by 41.1\% under HN-Cap and 42.9\% under Conflict. These margins all exceed its 31.8\% gain over UniVTG on clean input. Under Conflict, Suff.@2 also increases from 11.9 to 20.8, an 8.9-point gain.

Across all methods, incorrect-evidence perturbations cause a 24.9\% relative drop, nearly twice the 12.8\% drop under missing-source perturbations. Remaining modalities can partly compensate for a missing source, whereas plausible but incorrect evidence can redirect localization toward an incorrect moment. Robust localization therefore requires rejecting unreliable evidence, not merely retrieving more evidence.

\begin{table*}[t]
\centering
\small
\begin{tabular}{@{}p{0.34\textwidth}@{}*{4}{C{0.15\textwidth}@{}}}
\toprule
Setting & Cand. R & Sel. dur. & Final R1 & Cached time \\
& @0.3 $\uparrow$ & (\%) $\downarrow$ & @0.5 $\uparrow$ &
(s/query) $\downarrow$ \\
\midrule
\multicolumn{5}{@{}l@{}}{\emph{A. Planner comparison under Top-5}} \\
Uniform sampling & 31.8 & 11.6 & 11.9 & 0.54 \\
Text-only planner & 49.2 & 10.8 & 15.4 & 0.49 \\
Visual-only planner & 55.6 & 11.0 & 16.6 & 0.48 \\
MomentSeeker Top-5 \cite{yuan2025momentseeker} & 61.9 & 12.3 & 18.2 & 0.69 \\
EP-LVMQA w/o confidence & 68.4 & 10.9 & 20.9 & 0.53 \\
\textbf{EP-LVMQA Top-5 (default)} & \textbf{80.6} & 10.8 & \textbf{23.8} & 0.54 \\
\midrule
\multicolumn{5}{@{}l@{}}{\emph{B. Candidate-budget and oracle analysis}} \\
EP-LVMQA Top-3 & 71.1 & 6.5 & 21.4 & 0.40 \\
EP-LVMQA Top-10 & 88.9 & 20.2 & 24.7 & 0.83 \\
Oracle selection$^{\dagger}$ & 100.0 & -- & 31.2 & -- \\
\bottomrule
\end{tabular}
\caption{Planner and candidate-budget analysis on MedVidQA videos over 15\,min. Block A fixes Top-5; Block B varies $K$. Cand. R@0.3 is candidate recall; Sel. dur. is candidate-union/video duration. Cached time excludes offline extraction. $^{\dagger}$Oracle changes selection only; the fixed localizer receives no ground-truth boundaries.}
\label{tab:planner}
\end{table*}

Table~\ref{tab:planner} separates planner quality from candidate budget. Under the common Top-5 setting, EP-LVMQA achieved 80.6 candidate recall and 23.8 final $\Rone@0.5$. Relative to the same planner without confidence inputs, these scores increased by 12.2 and 2.9 points at nearly unchanged selected duration (10.8\% versus 10.9\%) and cached time (0.54 versus 0.53\,s/query). EP-LVMQA also exceeded the MomentSeeker Top-5 baseline by 18.7 candidate-recall points and 5.6 final localization points while selecting less video. The improvement therefore reflects better candidate selection rather than a larger evidence budget.

The budget sweep identifies Top-5 as the operating-point knee. Increasing $K$ from 3 to 5 added 9.5 candidate-recall points and 2.4 final $\Rone@0.5$ points for 4.3 additional duration points. Increasing $K$ from 5 to 10 added a similar 8.3 recall points but only 0.9 final points, while selected duration rose from 10.8\% to 20.2\% and cached time from 0.54 to 0.83\,s/query. Top-10 nevertheless remained 6.5 points below oracle selection, showing that increased candidate coverage alone cannot eliminate errors in candidate ranking and boundary refinement.

\subsection{Ablation and Human-Audited Evidence}
\begin{table*}[!t]
\centering
\small
\begin{tabular}{@{}p{0.19\textwidth}@{}*{7}{C{0.035\textwidth}@{}}*{7}{C{0.079\textwidth}@{}}}
\toprule
\multirow{2}{*}{Configuration} & \multirow{2}{*}{TL} & \multirow{2}{*}{Card} & \multirow{2}{*}{Src} & \multirow{2}{*}{Rtr} & \multirow{2}{*}{Cue} & \multirow{2}{*}{Ctx} & \multirow{2}{*}{Prior} & \multicolumn{2}{c}{R1} & \multirow{2}{*}{mIoU} & \multirow{2}{*}{Conflict} & \multirow{2}{*}{Suff.@2} & \multirow{2}{*}{Nec.@2} & \multirow{2}{*}{Ver.@2} \\
\cmidrule(lr){9-10}
& & & & & & & & @0.5 & @0.7 & & & & & \\
\midrule
\multicolumn{15}{@{}l}{\textit{Stage-level references}} \\
Standard MR reference & -- & -- & -- & -- & -- & -- & -- & 19.8 & 10.1 & 26.0 & 14.4 & 14.8 & 13.2 & 12.6 \\
Coarse planner only & -- & -- & -- & -- & -- & -- & -- & 20.5 & 8.7 & 27.0 & 15.8 & 17.6 & 15.4 & 14.7 \\
\midrule
\multicolumn{15}{@{}l}{\textit{Component ablations}} \\
$-$TL & $\times$ & $\checkmark$ & $\times$ & $\times$ & $\checkmark$ & $\checkmark$ & $\checkmark$ & 22.3 & 11.5 & 28.5 & 15.9 & 18.2 & 16.3 & 15.6 \\
$-$Card & $\checkmark$ & $\times$ & $\times$ & $\times$ & $\times$ & $\times$ & $\times$ & 21.7 & 10.9 & 27.7 & 16.5 & 16.9 & 15.1 & 14.3 \\
$-$Source fields & $\checkmark$ & $\checkmark$ & $\times$ & $\times$ & $\checkmark$ & $\checkmark$ & $\checkmark$ & 23.1 & 12.1 & 29.1 & 17.4 & 18.8 & 16.9 & 16.1 \\
$-$Confidence/router & $\checkmark$ & $\checkmark$ & $\checkmark$ & $\times$ & $\checkmark$ & $\checkmark$ & $\checkmark$ & 23.7 & 12.6 & 29.7 & 18.1 & 20.0 & 18.0 & 17.3 \\
$-$Cue & $\checkmark$ & $\checkmark$ & $\checkmark$ & $\checkmark$ & $\times$ & $\checkmark$ & $\checkmark$ & 24.2 & 13.0 & 30.1 & 19.6 & 21.6 & 19.5 & 18.8 \\
$-$Context & $\checkmark$ & $\checkmark$ & $\checkmark$ & $\checkmark$ & $\checkmark$ & $\times$ & $\checkmark$ & 24.6 & 13.3 & 30.4 & 20.3 & 22.3 & 20.4 & 19.4 \\
$-$Prior & $\checkmark$ & $\checkmark$ & $\checkmark$ & $\checkmark$ & $\checkmark$ & $\checkmark$ & $\times$ & 25.0 & 12.7 & 30.7 & 21.0 & 23.2 & 21.1 & 20.4 \\
\textbf{Full EP-LVMQA} & $\checkmark$ & $\checkmark$ & $\checkmark$ & $\checkmark$ & $\checkmark$ & $\checkmark$ & $\checkmark$ & \textbf{26.1} & \textbf{14.3} & \textbf{31.8} & \textbf{22.0} & \textbf{25.4} & \textbf{23.2} & \textbf{22.6} \\
\bottomrule
\end{tabular}
\caption{Stage-level references and component ablations on MedVidQA. Checks/crosses mark enabled/removed components or downstream unavailability. TL is the evidence timeline; Src contains modality evidence, masks, and confidence; Rtr is the source router; Conflict is R1@0.5 under ASR--caption conflict. Suff.@2 and Ver.@2 are primary outcomes; Nec.@2 is auxiliary. Audit values use the common blinded set and strict-majority rule; all variants share the split, encoders, Top-5 budget, backbone, optimizer, and evaluation.}
\label{tab:ablation}
\end{table*}

The card-mediated intervention affects the two primary audit outcomes more strongly than conventional overlap (Table~\ref{tab:ablation}). Removing it reduces $\Rone@0.5$ by 4.4 points, but reduces Suff.@2 and Ver.@2 by 8.5 and 8.3 points. Across the seven removals, the mean audit decline is roughly twice the corresponding $\Rone@0.5$ decline. Because removing the card also removes its dependent fields, this row estimates the whole planner-to-localizer evidence contract rather than the serialized container alone.

The stage-level references further separate evidence capture from temporal precision. Relative to standard moment retrieval, the coarse planner raises Suff.@2 from 14.8 to 17.6 but lowers $\Rone@0.7$ from 10.1 to 8.7. A coarse window can therefore contain answer-bearing evidence while remaining a poor temporal explanation. Fine localization is needed to convert semantic coverage into a sharply scoped moment.

Source identity accounts for much of the conflict result. Removing source fields lowers conflict $\Rone@0.5$ by 4.6 points and Suff.@2 by 6.6 points. Removing the prior has a larger effect at $\Rone@0.7$ than at $\Rone@0.5$ (1.6 versus 1.1 points), which is consistent with boundary sharpening. Cues and context produce smaller gains across all criteria. These effects are not additive: removing the timeline also disables downstream source fields and routing, so the $-$TL row measures the whole source-preservation pathway rather than the marginal value of timeline construction.

The ablations also form a consistent functional hierarchy. Removing the card produces the largest joint loss in localization and both primary audit outcomes. Removing source fields or the confidence/router pair most strongly affects conflict robustness, whereas the prior has a disproportionate effect at the stricter IoU threshold. Cue and context removal causes smaller but aligned declines. This ordering links the interface to evidence transfer, source-aware fields to reliability control, and the prior to boundary precision. Because several interventions disable dependent fields, the table supports pathway-level attribution rather than additive component effects.

\section{Conclusion}
\begin{figure}[t]
\centering
\includegraphics[width=\columnwidth]{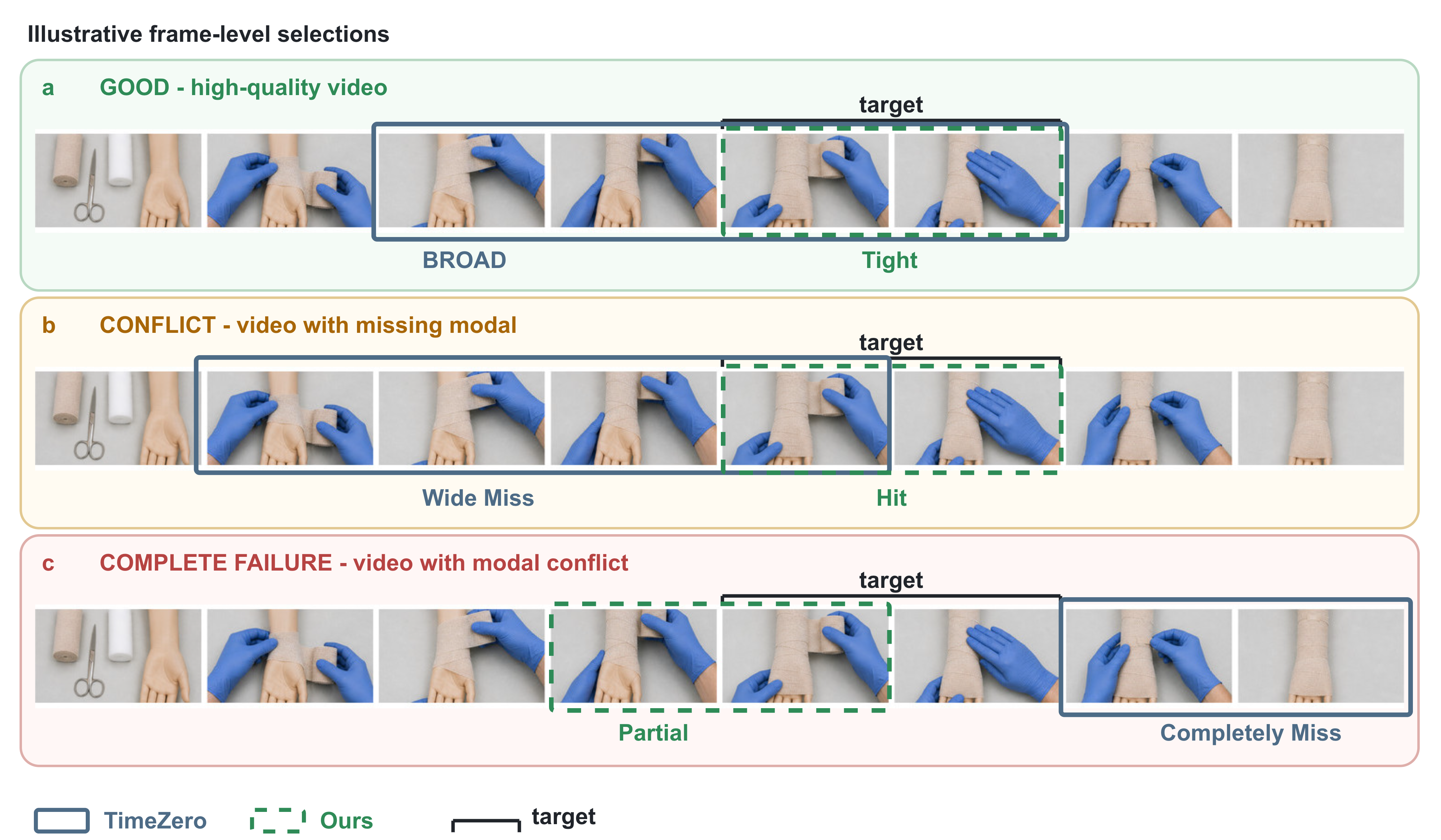}
\caption{Frame-level comparisons under complete, missing, and conflicting evidence. EP-LVMQA is tighter or more resilient than TimeZero, retaining target overlap with incomplete or inconsistent evidence.}
\label{fig:qualitative}
\end{figure}

EP-LVMQA couples an uncertainty-aware multimodal timeline, a coarse planner, and a source-routed fine localizer through a compact evidence card. On MedVidQA, it improves temporal localization, perturbation robustness, and audited evidence quality while remaining competitive on external benchmarks. Figure~\ref{fig:qualitative} illustrates tighter or more resilient selections under complete, missing, and conflicting evidence. Nevertheless, absolute audit rates remain too low for dependable human-facing deployment.

\clearpage
\bibliography{ep_lvmqa_refs}

@inproceedings{jang2017tgifqa,
  author    = {Jang, Yunseok and Song, Yale and Yu, Youngjae and Kim, Youngjin and Kim, Gunhee},
  title     = {{TGIF-QA}: Toward Spatio-Temporal Reasoning in Visual Question Answering},
  booktitle = {Proceedings of the IEEE Conference on Computer Vision and Pattern Recognition},
  pages     = {2758--2766},
  year      = {2017}
}

@inproceedings{lei2018tvqa,
  author    = {Lei, Jie and Yu, Licheng and Bansal, Mohit and Berg, Tamara L.},
  title     = {{TVQA}: Localized, Compositional Video Question Answering},
  booktitle = {Proceedings of the 2018 Conference on Empirical Methods in Natural Language Processing},
  pages     = {1369--1379},
  year      = {2018}
}

@inproceedings{xiao2024trust,
  author    = {Xiao, Junbin and Yao, Angela and Li, Yicong and Chua, Tat-Seng},
  title     = {Can I Trust Your Answer? Visually Grounded Video Question Answering},
  booktitle = {Proceedings of the IEEE/CVF Conference on Computer Vision and Pattern Recognition},
  pages     = {13204--13214},
  year      = {2024}
}

@inproceedings{di2024groundvqa,
  author    = {Di, Shangzhe and Xie, Weidi},
  title     = {Grounded Question-Answering in Long Egocentric Videos},
  booktitle = {Proceedings of the IEEE/CVF Conference on Computer Vision and Pattern Recognition},
  pages     = {12934--12943},
  year      = {2024}
}

@inproceedings{gao2017tall,
  author    = {Gao, Jiyang and Sun, Chen and Yang, Zhenheng and Nevatia, Ram},
  title     = {{TALL}: Temporal Activity Localization via Language Query},
  booktitle = {Proceedings of the IEEE International Conference on Computer Vision},
  pages     = {5267--5275},
  year      = {2017}
}

@inproceedings{lei2021qvhighlights,
  author    = {Lei, Jie and Berg, Tamara L. and Bansal, Mohit},
  title     = {Detecting Moments and Highlights in Videos via Natural Language Queries},
  booktitle = {Advances in Neural Information Processing Systems},
  volume    = {34},
  pages     = {11846--11858},
  year      = {2021}
}

@inproceedings{moon2023qddetr,
  author    = {Moon, WonJun and Hyun, Sangeek and Park, SangUk and Park, Dongchan and Heo, Jae-Pil},
  title     = {Query-Dependent Video Representation for Moment Retrieval and Highlight Detection},
  booktitle = {Proceedings of the IEEE/CVF Conference on Computer Vision and Pattern Recognition},
  pages     = {23023--23033},
  year      = {2023}
}

@article{moon2023cgdetr,
  author  = {Moon, WonJun and Hyun, Sangeek and Lee, SuBeen and Heo, Jae-Pil},
  title   = {Correlation-Guided Query-Dependency Calibration for Video Temporal Grounding},
  journal = {arXiv preprint arXiv:2311.08835},
  year    = {2023}
}

@inproceedings{lin2023univtg,
  author    = {Lin, Kevin Qinghong and Zhang, Pengchuan and Chen, Joya and Pramanick, Shraman and Gao, Difei and Wang, Alex Jinpeng and Yan, Rui and Shou, Mike Zheng},
  title     = {{UniVTG}: Towards Unified Video-Language Temporal Grounding},
  booktitle = {Proceedings of the IEEE/CVF International Conference on Computer Vision},
  pages     = {2794--2804},
  year      = {2023}
}

@inproceedings{hou2023cone,
  author    = {Hou, Zhijian and Zhong, Wanjun and Ji, Lei and Gao, Difei and Yan, Kun and Chan, Wing-Kwong and Ngo, Chong-Wah and Shou, Mike Zheng and Duan, Nan},
  title     = {{CONE}: An Efficient {CO}arse-to-fi{NE} Alignment Framework for Long Video Temporal Grounding},
  booktitle = {Proceedings of the 61st Annual Meeting of the Association for Computational Linguistics (Volume 1: Long Papers)},
  pages     = {8013--8028},
  year      = {2023},
  doi       = {10.18653/v1/2023.acl-long.445}
}

@article{hou2023groundnlq,
  author  = {Hou, Zhijian and Ji, Lei and Gao, Difei and Zhong, Wanjun and Yan, Kun and Li, Chao and Chan, Wing-Kwong and Ngo, Chong-Wah and Duan, Nan and Shou, Mike Zheng},
  title   = {{GroundNLQ} @ {Ego4D} Natural Language Queries Challenge 2023},
  journal = {arXiv preprint arXiv:2306.15255},
  year    = {2023}
}

@inproceedings{pan2023scanning,
  author    = {Pan, Yulin and He, Xiangteng and Gong, Biao and Lv, Yiliang and Shen, Yujun and Peng, Yuxin and Zhao, Deli},
  title     = {Scanning Only Once: An End-to-End Framework for Fast Temporal Grounding in Long Videos},
  booktitle = {Proceedings of the IEEE/CVF International Conference on Computer Vision},
  pages     = {13721--13731},
  year      = {2023}
}

@inproceedings{barrios2023guidance,
  author    = {Barrios, Wayner and Soldan, Mattia and Ceballos-Arroyo, Alberto Mario and Heilbron, Fabian Caba and Ghanem, Bernard},
  title     = {Localizing Moments in Long Video via Multimodal Guidance},
  booktitle = {Proceedings of the IEEE/CVF International Conference on Computer Vision},
  pages     = {13667--13678},
  year      = {2023}
}

@inproceedings{mu2024snag,
  author    = {Mu, Fangzhou and Mo, Sicheng and Li, Yin},
  title     = {{SnAG}: Scalable and Accurate Video Grounding},
  booktitle = {Proceedings of the IEEE/CVF Conference on Computer Vision and Pattern Recognition},
  pages     = {18930--18940},
  year      = {2024}
}

@article{gupta2023medvidqa,
  author  = {Gupta, Deepak and Attal, Kush and Demner-Fushman, Dina},
  title   = {A Dataset for Medical Instructional Video Classification and Question Answering},
  journal = {Scientific Data},
  volume  = {10},
  number  = {1},
  pages   = {158},
  year    = {2023}
}

@inproceedings{zhang2020vslnet,
  author    = {Zhang, Hao and Sun, Aixin and Jing, Wei and Zhou, Joey Tianyi},
  title     = {Span-Based Localizing Network for Natural Language Video Localization},
  booktitle = {Proceedings of the 58th Annual Meeting of the Association for Computational Linguistics},
  pages     = {6543--6554},
  year      = {2020}
}

@inproceedings{grauman2022ego4d,
  author    = {Grauman, Kristen and Westbury, Andrew and Byrne, Eugene and others},
  title     = {{Ego4D}: Around the World in 3,000 Hours of Egocentric Video},
  booktitle = {Proceedings of the IEEE/CVF Conference on Computer Vision and Pattern Recognition},
  pages     = {18995--19012},
  year      = {2022}
}

@inproceedings{yuan2025momentseeker,
  author    = {Yuan, Huaying and Ni, Jian and Liu, Zheng and Wang, Yueze and Zhou, Junjie and Liang, Zhengyang and Zhao, Bo and Cao, Zhao and Wen, Ji-Rong and Dou, Zhicheng},
  title     = {{MomentSeeker}: A Task-Oriented Benchmark for Long-Video Moment Retrieval},
  booktitle = {Advances in Neural Information Processing Systems},
  volume    = {38},
  year      = {2025}
}

@inproceedings{huang2024vtimellm,
  author    = {Huang, Bin and Wang, Xin and Chen, Hong and Song, Zihan and Zhu, Wenwu},
  title     = {{VTimeLLM}: Empower {LLM} to Grasp Video Moments},
  booktitle = {Proceedings of the IEEE/CVF Conference on Computer Vision and Pattern Recognition},
  pages     = {14271--14280},
  year      = {2024}
}

@inproceedings{wang2025grounded,
  author    = {Wang, Haibo and Xu, Zhiyang and Cheng, Yu and Diao, Shizhe and Zhou, Yufan and Cao, Yixin and Wang, Qifan and Ge, Weifeng and Huang, Lifu},
  title     = {Grounded-VideoLLM: Sharpening Fine-Grained Temporal Grounding in Video Large Language Models},
  booktitle = {Findings of the Association for Computational Linguistics: EMNLP 2025},
  pages     = {959--975},
  year      = {2025}
}

@article{wang2025timezero,
  author  = {Wang, Ye and Xu, Boshen and Yue, Zihao and Xiao, Zihan and Wang, Ziheng and Zhang, Liang and Yang, Dingyi and Wang, Wenxuan and Jin, Qin},
  title   = {{TimeZero}: Temporal Video Grounding with Reasoning-Guided {LVLM}},
  journal = {arXiv preprint arXiv:2503.13377v1},
  year    = {2025},
  eprint  = {2503.13377v1},
  archiveprefix = {arXiv},
  url     = {https://arxiv.org/abs/2503.13377v1}
}

@inproceedings{wang2025causalvtg,
  author    = {Wang, Qiyi and Chen, Senda and Shen, Ying},
  title     = {{CausalVTG}: Towards Robust Video Temporal Grounding via Causal Inference},
  booktitle = {Advances in Neural Information Processing Systems},
  volume    = {38},
  year      = {2025}
}

@article{lee2026srf,
  author  = {Lee, YuEun and Kim, Jung Uk},
  title   = {See, Rank, and Filter: Important Word-Aware Clip Filtering via Scene Understanding for Moment Retrieval and Highlight Detection},
  journal = {Proceedings of the AAAI Conference on Artificial Intelligence},
  volume  = {40},
  number  = {8},
  pages   = {5936--5944},
  year    = {2026},
  doi     = {10.1609/aaai.v40i8.37516}
}

@inproceedings{feichtenhofer2019slowfast,
  author    = {Feichtenhofer, Christoph and Fan, Haoqi and Malik, Jitendra and He, Kaiming},
  title     = {SlowFast Networks for Video Recognition},
  booktitle = {Proceedings of the IEEE/CVF International Conference on Computer Vision},
  pages     = {6202--6211},
  year      = {2019}
}

@inproceedings{radford2021clip,
  author    = {Radford, Alec and Kim, Jong Wook and Hallacy, Chris and Ramesh, Aditya and Goh, Gabriel and Agarwal, Sandhini and Sastry, Girish and Askell, Amanda and Mishkin, Pamela and Clark, Jack and Krueger, Gretchen and Sutskever, Ilya},
  title     = {Learning Transferable Visual Models from Natural Language Supervision},
  booktitle = {Proceedings of the 38th International Conference on Machine Learning},
  series    = {Proceedings of Machine Learning Research},
  volume    = {139},
  publisher = {PMLR},
  pages     = {8748--8763},
  year      = {2021}
}

@inproceedings{radford2023whisper,
  author    = {Radford, Alec and Kim, Jong Wook and Xu, Tao and Brockman, Greg and McLeavey, Christine and Sutskever, Ilya},
  title     = {Robust Speech Recognition via Large-Scale Weak Supervision},
  booktitle = {Proceedings of the 40th International Conference on Machine Learning},
  series    = {Proceedings of Machine Learning Research},
  volume    = {202},
  publisher = {PMLR},
  pages     = {28492--28518},
  year      = {2023}
}

@article{du2020ppocr,
  author  = {Du, Yuning and Li, Chenxia and Guo, Ruoyu and Yin, Xiaoting and Liu, Weiwei and Zhou, Jun and Bai, Yifan and Yu, Zilin and Yang, Yehua and Dang, Qingqing and Wang, Haoshuang},
  title   = {{PP-OCR}: A Practical Ultra Lightweight {OCR} System},
  journal = {arXiv preprint arXiv:2009.09941},
  year    = {2020}
}

@article{li2024llavaonevision,
  author  = {Li, Bo and Zhang, Yuanhan and Guo, Dong and Zhang, Renrui and Li, Feng and Zhang, Hao and Zhang, Kaichen and Zhang, Peiyuan and Li, Yanwei and Liu, Ziwei and Li, Chunyuan},
  title   = {{LLaVA-OneVision}: Easy Visual Task Transfer},
  journal = {Transactions on Machine Learning Research},
  year    = {2025},
  url     = {https://openreview.net/forum?id=zKv8qULV6n}
}

@inproceedings{alayrac2020selfsupervised,
  author    = {Alayrac, Jean-Baptiste and Recasens, Adria and Schneider, Rosalia and Arandjelovi{\'c}, Oisin and Ramapuram, Jason and De Fauw, Jeffrey and Smaira, Lucas and Dieleman, Sander and Zisserman, Andrew},
  title     = {Self-Supervised {MultiModal} Versatile Networks},
  booktitle = {Advances in Neural Information Processing Systems},
  volume    = {33},
  pages     = {25--37},
  year      = {2020}
}

@inproceedings{bain2021frozen,
  author    = {Bain, Max and Nagrani, Arsha and Varol, G{\"u}l and Zisserman, Andrew},
  title     = {Frozen in Time: A Joint Video and Image Encoder for End-to-End Retrieval},
  booktitle = {Proceedings of the IEEE/CVF International Conference on Computer Vision},
  pages     = {1728--1738},
  year      = {2021}
}

@inproceedings{carbonell1998mmr,
  author    = {Carbonell, Jaime and Goldstein, Jade},
  title     = {The Use of {MMR}, Diversity-Based Reranking for Reordering Documents and Producing Summaries},
  booktitle = {Proceedings of the 21st Annual International ACM SIGIR Conference on Research and Development in Information Retrieval},
  pages     = {335--336},
  year      = {1998}
}

@inproceedings{ghosh2019excl,
  author    = {Ghosh, Soham and Agarwal, Aakanksha and Parekh, Zarana and Hauptmann, Alexander},
  title     = {{ExCL}: Extractive Clip Localization Using Natural Language Descriptions},
  booktitle = {Proceedings of the 2019 Conference of the North American Chapter of the Association for Computational Linguistics: Human Language Technologies, Volume 1},
  pages     = {1984--1990},
  year      = {2019}
}

@inproceedings{guo2017calibration,
  author    = {Guo, Chuan and Pleiss, Geoff and Sun, Yu and Weinberger, Kilian Q.},
  title     = {On Calibration of Modern Neural Networks},
  booktitle = {Proceedings of the 34th International Conference on Machine Learning},
  series    = {Proceedings of Machine Learning Research},
  volume    = {70},
  pages     = {1321--1330},
  publisher = {PMLR},
  year      = {2017}
}

@inproceedings{hendricks2017localizing,
  author    = {Hendricks, Lisa Anne and Wang, Oliver and Shechtman, Eli and Sivic, Josef and Darrell, Trevor and Russell, Bryan},
  title     = {Localizing Moments in Video with Natural Language},
  booktitle = {Proceedings of the IEEE International Conference on Computer Vision},
  pages     = {5803--5812},
  year      = {2017}
}

@inproceedings{liu2022umt,
  author    = {Liu, Ye and Li, Siyuan and Wu, Yang and Chen, Chang-Wen and Shan, Ying and Qie, Xian-Sheng},
  title     = {{UMT}: Unified Multi-Modal Transformers for Joint Video Moment Retrieval and Highlight Detection},
  booktitle = {Proceedings of the IEEE/CVF Conference on Computer Vision and Pattern Recognition},
  pages     = {3042--3051},
  year      = {2022}
}

@inproceedings{ma2022missing,
  author    = {Ma, Mengmeng and Ren, Jian and Zhao, Long and Testuggine, Davide and Peng, Xuyang},
  title     = {Are Multimodal Transformers Robust to Missing Modality?},
  booktitle = {Proceedings of the IEEE/CVF Conference on Computer Vision and Pattern Recognition},
  pages     = {18177--18186},
  year      = {2022}
}

@inproceedings{miech2019howto100m,
  author    = {Miech, Antoine and Zhukov, Dmitry and Alayrac, Jean-Baptiste and Tapaswi, Makarand and Laptev, Ivan and Sivic, Josef},
  title     = {{HowTo100M}: Learning a Text-Video Embedding by Watching Hundred Million Narrated Video Clips},
  booktitle = {Proceedings of the IEEE/CVF International Conference on Computer Vision},
  pages     = {2630--2640},
  year      = {2019}
}

@article{neverova2016moddrop,
  author  = {Neverova, Natalia and Wolf, Christian and Taylor, Graham W. and Nebout, Florian},
  title   = {{ModDrop}: Adaptive Multi-Modal Gesture Recognition},
  journal = {IEEE Transactions on Pattern Analysis and Machine Intelligence},
  volume  = {38},
  number  = {8},
  pages   = {1692--1706},
  year    = {2016}
}

@article{regneri2013grounding,
  author  = {Regneri, Michaela and Rohrbach, Marcus and Wetzel, Dominikus and Thater, Stefan and Schiele, Bernt and Pinkal, Manfred},
  title   = {Grounding Action Descriptions in Videos},
  journal = {Transactions of the Association for Computational Linguistics},
  volume  = {1},
  pages   = {25--36},
  year    = {2013}
}

@inproceedings{soldan2022mad,
  author    = {Soldan, Mattia and Pardo, Andres and Alc{\'a}zar, Juan Luis and Heilbron, Fabian Caba and Zhao, Chen and Giancola, Silvio and Ghanem, Bernard},
  title     = {{MAD}: A Scalable Dataset for Language Grounding in Videos from Movie Audio Descriptions},
  booktitle = {Proceedings of the IEEE/CVF Conference on Computer Vision and Pattern Recognition},
  pages     = {5026--5035},
  year      = {2022}
}

@inproceedings{zhang2020t2tan,
  author    = {Zhang, Sijia and Peng, Haohan and Fu, Jianlong and Luo, Jiebo},
  title     = {Learning 2D Temporal Adjacent Networks for Moment Localization with Natural Language},
  booktitle = {Proceedings of the AAAI Conference on Artificial Intelligence},
  volume    = {34(7)},
  pages     = {12870--12877},
  year      = {2020}
}
\end{document}